\documentclass[9pt, conference]{IEEEtran}

\usepackage{hyperref}
\usepackage{tabularx}
\usepackage{amsmath}
\usepackage{multirow}
\usepackage{listings}
\usepackage{graphicx}
\usepackage{booktabs}
\usepackage{algorithm}
\usepackage{algorithmicx}
\usepackage{algpseudocode}
\usepackage{multirow}

\begin{document}

\title{NASH: Neural Architecture Search for Hardware-Optimized Machine Learning Models}

\author{\IEEEauthorblockN {Mengfei Ji, Yuchun Chang, Baolin Zhang and
Zaid Al-Ars}
\IEEEauthorblockA{Jilin University\\
Changchun, Jilin, China\\
Delft University of Technology\\
Delft, The Netherlands\\ 
Email: jimengfei007@outlook.com\\
}
}

\maketitle

\begin{abstract}

As machine learning (ML) algorithms get deployed in an ever-increasing number of applications, these algorithms need to achieve better trade-offs between high accuracy, high throughput and low latency. This paper introduces NASH, a novel approach that applies neural architecture search to machine learning hardware. Using NASH, hardware designs can achieve not only high throughput and low latency but also superior accuracy performance. We present four versions of the NASH strategy in this paper, all of which show higher accuracy than the original models. The strategy can be applied to various convolutional neural networks, selecting specific model operations among many to guide the training process toward higher accuracy. Experimental results show that applying NASH on ResNet18 or ResNet34 achieves a top 1 accuracy increase of up to 3.1\% and a top 5 accuracy increase of up to 2.2\% compared to the non-NASH version when tested on the ImageNet data set. We also integrated this approach into the FINN hardware model synthesis tool to automate the application of our approach and the generation of the hardware model.
Results show that using FINN can achieve a maximum throughput of 324.5 fps. In addition, NASH models can also result in a better trade-off between accuracy and hardware resource utilization. The accuracy-hardware (HW) Pareto curve shows that the models with the four NASH versions represent the best trade-offs achieving the highest accuracy for a given HW utilization. The code for our implementation is open-source and publicly available on GitHub at https://github.com/MFJI/NASH.

\end{abstract}


\IEEEpeerreviewmaketitle

\section{Introduction}

The rise of ML has impacted many aspects of our daily lives from web searches to autonomous driving \cite{WOS:000754451800010}\cite{WOS:000809323500001}\cite{WOS:000735517000006}. In many situations, the application imposes strict demands on ML model behavior, requiring high accuracy, high throughput, low latency, low HW utilization, and/or low power. Designing custom hardware implementations of these models can help achieve many of these requirements.
However, hardware implementations require the use of low-bit width data, which leads to a decrease in the accuracy of the model, thereby limiting the applicability of such solutions in practice.
To ensure the wide applicability of custom hardware ML model implementations, we need to investigate advanced methods to improve the accuracy of low-bit-width models (e.g., binarized models). 

Existing solutions focus on software-based improvement of these models using multi-weight binary models \cite{bengio2013estimating}, or linear combinations of binary weights \cite{WOS:000650971302012}, or improving gradient paths\cite{DBLP:conf/bmvc/ZhuHLA22}. Another approach is to use neural architecture search (NAS) to increase the accuracy of the networks \cite{WOS:000626021408054}. Although these methods are able to improve model accuracy, they are fully software-based and cannot be used directly to improve the accuracy of hardware implementations. 

In this paper, we propose NASH, a hardware-based architecture search algorithm that optimizes the architecture of an ML model specifically for hardware implementations. The design uses the open-source hardware compiler FINN ~\cite{WOS:000455722100008}~\cite{WOS:000457149400003}, which is an efficient framework to explore deep neural network inference on FPGAs specifically for quantized neural networks. The implementation ensures high accuracy and low HW utilization of the generated model for the used hardware. Four versions of the NASH strategy are shown in this paper and are able to achieve the best trade-off between accuracy and HW resource utilization.

The contributions of this paper are as follows:

\begin{enumerate}
\itemsep=0pt
\item We propose an approach to perform an architecture search for ML models implemented in hardware.
\item We integrate this approach into the FINN hardware model synthesis tool to automate its application.
\item We demonstrate our methodology by building a hardware-optimized ResNet18 and ResNet34 that show up to 3.0\% higher accuracy and show a better trade-off between accuracy and hardware resource utilization.
\end{enumerate}  

The rest of this paper is organized as follows. Section \ref{Background} provides some background information, including the neural architecture search approach and the FINN automated convolution neural network inference hardware implementation tool. In Section \ref{NASH approach}, we present NASH, which extends the neural architecture search approach to hardware. Section \ref{Experimental Results} shows the experimental results of applying NASH to image classification models in hardware. Section \ref{conclusion} concludes the paper.

\section{Background}
\label{Background}
This section introduces NAS and the FINN framework that we use as building blocks for our approach.

\subsection{Neural architecture search}

Traditionally, neural networks are carefully designed by researchers based on expert knowledge in the field, which makes this process labor-intensive and time-consuming. As an alternative, NAS provides an automatic and efficient method for selecting the best architecture alternative for a specific neural network. NAS explores various network components (such as the convolution layer and the max pool layer), network topology (including the number of layers and channels used), and hyperparameters (such as the learning rate and regularization strength). The NAS method simultaneously searches the network structure and updates its weights. Networks designed using NAS show higher accuracy, compared to manually designed networks.

Various NAS approaches have been reported in recent years. For instance, ENAS \cite{WOS:000683379204022} performs well empirically since it shares parameters with child models. PNAS \cite{WOS:000594203000002} looks for structures in the structure space in the order of increasing complexity while concurrently learning a surrogate model to direct the search. However, this continually stacks a group of similar blocks in the model resulting in inefficiencies. Instead, FPNAS \cite{WOS:000548549201033} explores the entire network design to ensure block diversity, which is useful for increasing the efficiency of the networks.

Usually, using the NAS approach requires a high demand for computation resources. Thousands of GPU days \cite{WOS:000457843608090}\cite{WOS:000485292604099} are needed to search for the best architecture on large data sets like ImageNet. Continuous neural architecture search addresses this issue properly. Differentiable continuous neural architecture search \cite{liu2019darts}\cite{xie2018snas} makes it possible to search the model structure on a small data set and then adopt it to a larger data set. However, none of these approaches is focused on generating optimal model architectures for quantized neural networks.

NASB proposes the first binary network neural architecture search algorithm~\cite{WOS:000626021408054}. The architecture search algorithm is adjusted to include structures and operations suitable for hardware. NASB finds a specific NASB-convolutional cell as an optimal architecture for binarizing its full precision counterpart, where the NASB-convolutional cell can be a replacement for a binarized convolutional layer, block, group, and network. However, NASB does not provide an approach to optimize the HW implementation of the binary neural network.

In this paper, we base our work on the continuous neural architecture search algorithm, extending it to be applicable to hardware.

\subsection{FINN framework}
\label{FINN framework}
FINN is an open-source framework developed by AMD Xilinx for generating fast and scalable quantized convolution neural network inference accelerators on FPGAs \cite{WOS:000455722100008}\cite{WOS:000457149400003}. The input files of FINN are the Open Neural Network Exchange (ONNX) format exported by Brevitas \cite{brevitas}, which is a PyTorch library for neural network quantization, with a focus on quantization-aware training. FINN provides an end-to-end flow to generate heterogeneous streaming architectures for different network topologies. FINN is a flexible framework where both throughput and design resources can be considered. The number of SIMD lanes and the number of processing elements (PEs) determine how each layer of a CNN is folded in a Matrix-Vector-Threshold Unit (MVTU). The SIMD lanes are the corresponding synapses of the PEs, which are hardware representations of the neurons in a neural network and contain the weights as well as conduct the multiplication operation. While utilizing folding, the number of PEs selected must be a factor of all the neurons. The number of SIMDs follows the same logic.
FINN provides an easy approach to hardware design in many aspects. Previous research \cite{WOS:000735988700001} used FINN to generate a PointPillars Network for object detection based on a point cloud obtained by a LiDAR sensor. Another example is using FINN to generate a model for modulation classification for RadioML on RFSoC \cite{WOS:000878171600028}. There is also an extension of the FINN library for Long Short-Term Memory neural networks \cite{WOS:000460538500017}, which makes the automatic generation of recurrent neural networks possible. Because of its efficiency and flexibility, FINN is widely used in generating hardware implementations for many cutting-edge designs. In this paper, FINN is used as the tool to generate our networks for hardware implementations.

\section{NASH approach}
\label{NASH approach}
In this section, we introduce our NASH approach, describing the overall hardware architecture search process. Then, we discuss the structure of the model and a discussion of the operations that the model uses. After that, we introduce the training algorithm of the NASH approach, followed by the way we implement the hardware.

\subsection{Overall process}
\begin{figure}
    \centering
    \includegraphics [width=\columnwidth]{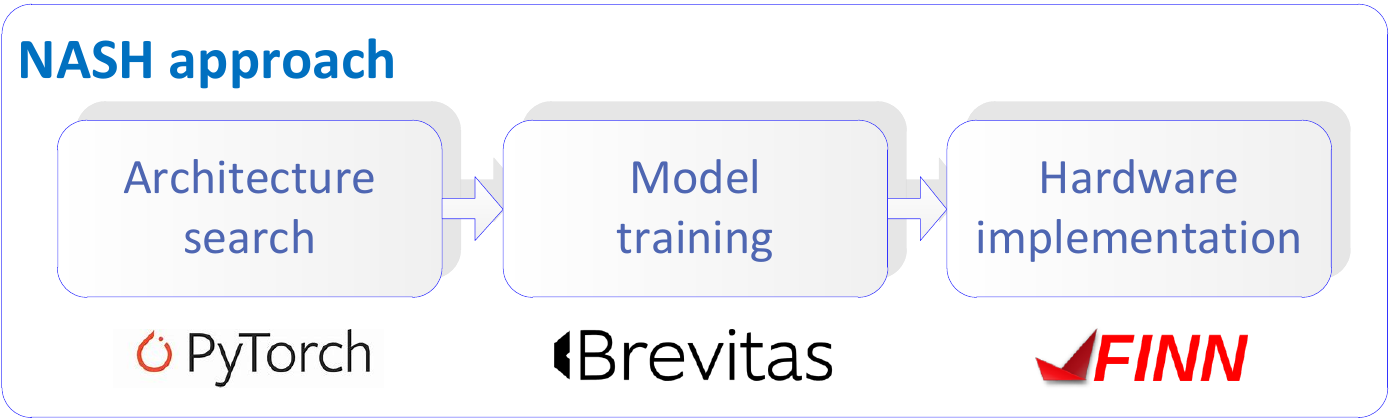}
    \caption{Overall process of the NASH approach}
    \label{fig3.1}
\end{figure}

Our NASH approach can be divided into 3 main stages: architecture search, model training, and hardware implementation, as shown in Figure \ref{fig3.1}. We use differentiable neural architecture search \cite{WOS:000626021408054} \cite{WOS:000457843608090} to develop the first two stages.
The architecture search algorithm is adjusted to include structures and operations suitable for hardware. In addition, the model training algorithm is adjusted to implement model quantization techniques. Finally, we use the FINN tool to implement the trained model in hardware.

\subsection{NASH structure}
\label{NASH structure}
\begin{figure*}
    \centering
    \includegraphics [width=1.5\columnwidth]{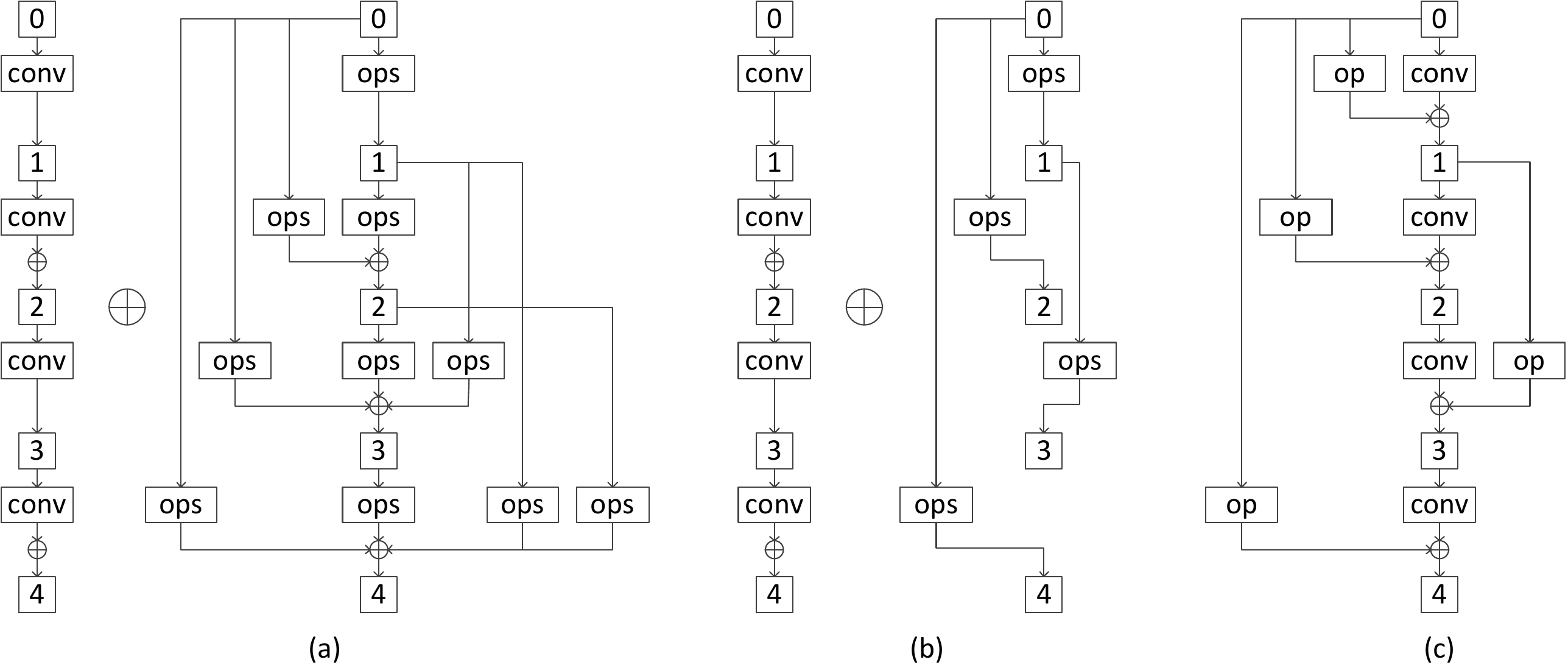}
    \caption{Connections of a NASH-convolutional cell}
    \label{fig3.2}
\end{figure*}

The NASH approach uses the NAS strategy to identify a convolutional cell to optimize the architecture of a network, such that it can be conveniently translated into a hardware implementation. NASH uses a number of NASH-convolutional cell alternatives such as a convolutional layer, block, group, and network. Figure \ref{fig3.2} illustrates the connections of a NASH-convolutional cell. In the figure, conv refers to a quantized convolutional layer and ops refers to a set of operations, among which only one operation is active during the model architecture search stage. $\oplus$ refers to the element-wise addition between the tensors of the two nodes with the same number.

Figure \ref{fig3.2}(a) illustrates the connections of a NASH-convolutional cell used in the training of the searching stage, which includes a backbone representing a standard convolutional group (left), and a NAS-convolutional cell representing a directed acyclic graph (right). The NAS-convolutional cell consists of five nodes, ten edges, and five operations for every edge (shown as "ops" in the figure), and it is designed to maintain the layer depth of the NASH-convolutional cell identical to that of its full precision equivalent, thereby avoiding any increase in latency during inference. The connections of the backbone are fixed, and there is no need to specify architecture parameters for it. During the training of the searching stage, the weights of the NASH-convolutional cell model and the architecture parameters of the NAS-convolutional cell can be updated alternately, and only one operation on every edge of the NAS-convolutional cell is sampled and active at each step to decrease memory requirements by deactivating inactive paths.

Figure \ref{fig3.2}(b) presents an example of the final architecture after finishing the training of the searching stage. In the NAS-convolutional cell, only one predecessor is retained for each node, and one operation is kept for each edge except for node 0. Figure \ref{fig3.2}(c) provides the combined representation of the backbone and the NAS-convolutional cell of Figure \ref{fig3.2}(b), outlining the output of each node (excluding node 0) in the NASH-convolutional cell. Each node is computed based on all of its predecessors:
\[{x^{(j)}} = o{p^{(i,j)}}({x^{(i)}}) \oplus con{v^{(i,j)}}({x^{(i)}})\]
where $x^{(i)}$ is the input node and $x^{(j)}$ is the output node.

\begin{figure}
    \centering
    \includegraphics [width=0.9\columnwidth]{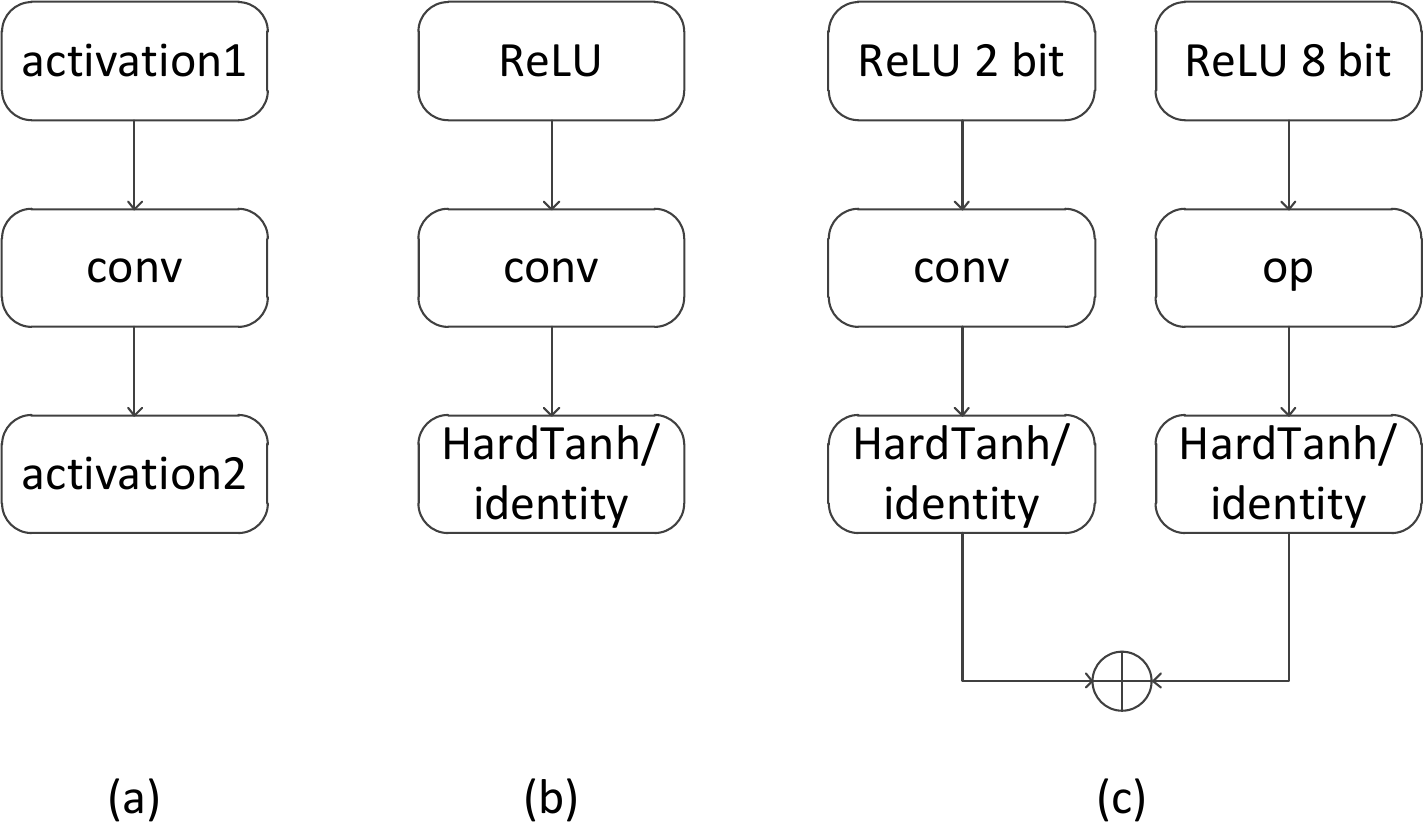}
    \caption{Detailed structure of the "conv" and "op" layers in Figure \ref{fig3.2}(c)}
    \label{fig3.3}
\end{figure}

After introducing the structure of the NASH-convolutional cell, we will look at the structure in the convolutional layers in the backbone of the NASH-convolutional cell. Because we are using FINN to generate our NASH models as a hardware implementation, and since FINN requires activation layers before every addition of the branches, we place an activation function before each one of these additions. This is shown as "activation2" in Figure \ref{fig3.3}(a). In addition, since an activation function is needed to quantize the input of the convolutional layer, we place an activation function at the beginning of each branch. This is shown as "activation1" in Figure \ref{fig3.3}(a). 

The two activation functions shown in Figure \ref{fig3.3}(a) can be implemented as one of the following functions identity, HardTanh, and ReLU activation. We can use an arbitrary number of bits for these functions with bit widths ranging from 1 bit up to 8 bits. In addition, since the structure of these three functions is different in hardware, we have to use ReLU for the function "activation1", while we can use identity or HardTanh for the function "activation2". The structure of the convolutional layer after selecting the activation function is shown in Figure \ref{fig3.3}(b).

The structure of the NAS layers is similar to the backbone layers, only replacing the convolutional function with the functions searched by the architecture search algorithm. In order to make sure the accuracy of the NASH models is high, we choose to use higher bit width activation (8 bit) for the NAS operations, because we have max pooling operations in NAS branches (a 2-bit input for max pooling operations does not give high accuracy). Therefore, the activation function ReLU uses different bit widths for the backbone layers (2 bit) and the NAS layers (8 bit) as shown in Figure \ref{fig3.3}(c).

\begin{figure}
    \centering
    \includegraphics [width=\columnwidth]{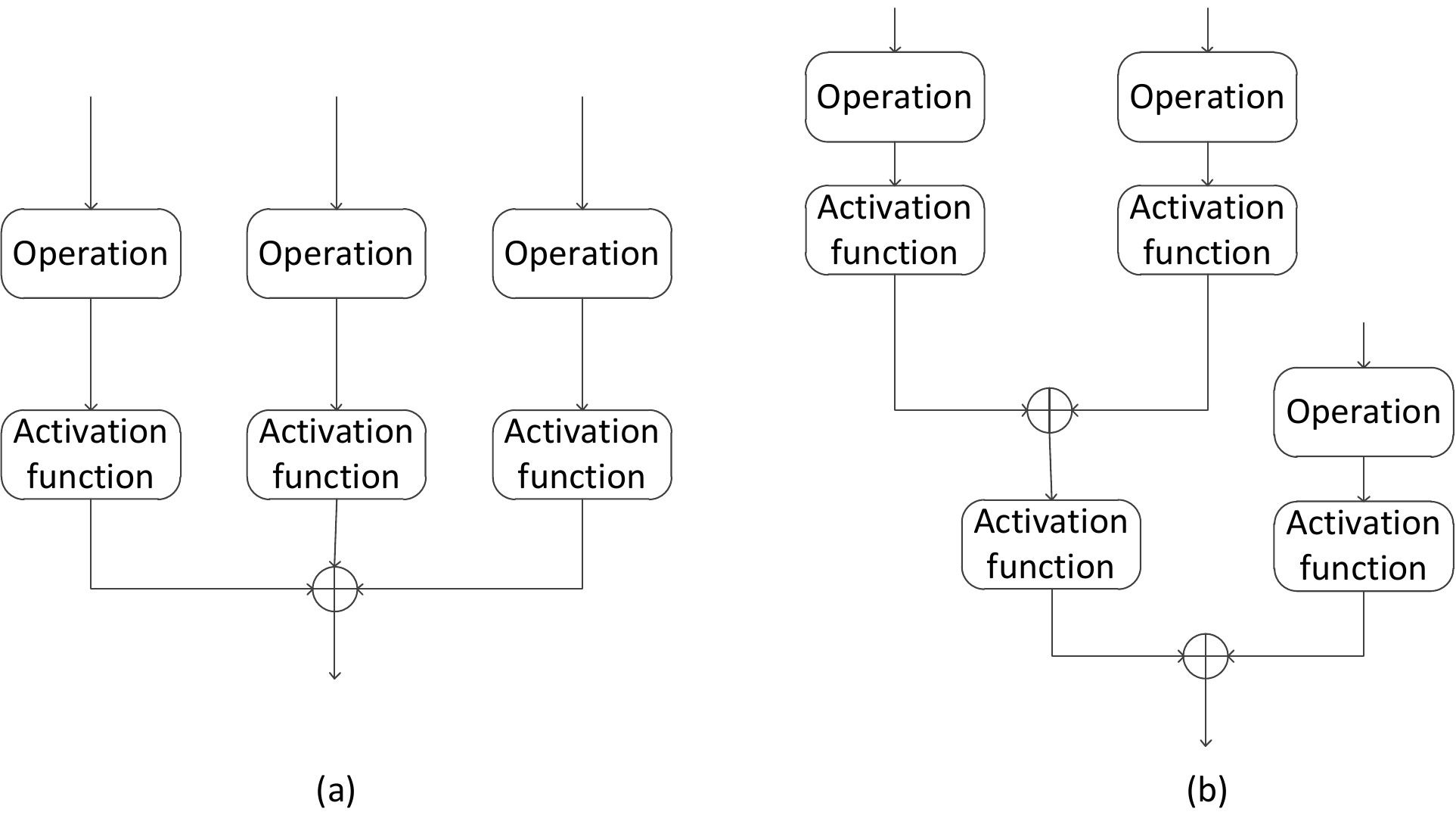}
    \caption{Parallel (a) and cascade (b) structure of multi-branch additions}
    \label{fig3.4}
\end{figure}

When combining more than two branches together, we need to generalize the addition operation in hardware. This is done by using a cascade structure instead of the parallel structure as shown in Figure \ref{fig3.4}. This facilitates the hardware generation of branch addition in FINN. 

\subsection{NASH operations}
\label{NASH operations}
There are multiple choices that we can use to represent the operations in NASH: zero operation, convolution operations, max pooling operation, and identity operation. We list the operation we use in the NASH strategy in Table \ref{tb3.1}.

\begin{table}[]
    \centering
    \caption{NASH operations used in the NAS-convolutional cell}
    \label{tb3.1}
    \begin{tabular}{ll}
    \hline
    Operation & Description\\
    \hline
    op0 & Zero\\
    \hline
    op1 & 3$\times$3 max pooling\\
    \hline
    op2 & Identity\\
    \hline
    op3 & 1$\times$1 convolution\\
    \hline
    op4 & 3$\times$3 convolution\\
    \hline
    op5 & 5$\times$5 convolution\\
    \hline
    \end{tabular}
\end{table}

 Among the operations in Table \ref{tb3.1}, op1 (the max pooling operation) is not hardware-friendly. Because max pooling operations cannot change the size between the input and output channel, an extra layer (the concatenate layer) needs to be used when the branch needs to change the size of the input layer to the output layer. However, the concatenate layer causes extra challenges when generating the hardware. Therefore, we choose to limit the usage of the max pooling operations when we need to change the output size.

\subsection{Training algorithm}
\label{Training algorithm}

The training algorithm includes the algorithms in the first two stages of the NASH approach, which are model architecture search and model training (Figure \ref{fig3.1}). We use PyTorch~\cite{paszke2017automatic} in the model architecture search stage and we use Brevitas in the model training stage. Brevitas is able to generate quantized models, which is important for our hardware implementation. The model architecture search stage uses a smaller data set (CIFAR-10) to train which saves compute resources, while the model training stage uses a larger data set (ImageNet). This is possible, since the model architecture searched on one data set is also suitable for another \cite{liu2019darts}. Performing the searching stage on a small data set rather than directly on the target data set can be regarded as a proxy task to find the optimized architecture model for the model training stage, which can enable a large search space and significantly accelerate the computation of the NASH strategy.

The goal of the searching stage is to get an optimized CNN architecture, which is done by using NAS to train a CNN model from scratch on a data set. The algorithm used in the model architecture search stage is listed as Algorithm \ref{a1}.

\begin{algorithm}
\caption{Model architecture search algorithm}\label{a1}
\begin{algorithmic}[1]
\Statex \textbf{Input:} Dataset D1
\Statex \textbf{Output:} CNN model M1
\For{$epoch = 1$ to $L$}
    \For{$batch = 1$ to $N$}
    \State 
    Randomly sample a mini-batch validation data from D1, freeze the model weights of model M1, and update its architecture parameters. 
        \State {Randomly sample a mini-batch training data from D1, freeze the architecture parameters of model M1, and update its model weights.}        
    \EndFor
\EndFor
\end{algorithmic}
\end{algorithm}

In the model architecture search stage of NASH, we are searching for the model architecture that is possible for hardware implementation. In this paper, we present four versions of algorithms to search for the best architecture of the model which is suitable to implement in hardware; these algorithms are referred to as NASH-v1 to NASH-v4. 

NASH-v1 replaces all 3$\times$3 max pooling operations (that change the output size) with a 1$\times$1 convolution after the basic model architecture search algorithm is completed. In NASH-v2, during the model architecture search, if the algorithm identifies that the branch needs to change the output channel size, then the 3$\times$3 max pooling layer is not used in this search. NASH-v3 directly discards any search results with unsuitable 3$\times$3 max pooling operations and searches for the second-best one until the result is acceptable. NASH-v4 is the same as the basic model architecture search algorithm, where the only difference is that we directly removed the 3$\times$3 max pooling operation from the list of alternative operations of NASH. After searching with NASH-v4, it is the case that more convolutional layers will be used in the searching branches, so in this way, more hardware resources are needed.

We evaluated the four versions of the architecture search algorithm to train the models and identify the model with the best accuracy. We also tested the accuracy of the models of all four versions in the subsequent model training stage, and verified that all four models have better accuracy than the original model without the NASH strategy. 

\begin{algorithm}
\caption{Model training algorithm}\label{a2}
\begin{algorithmic}[1]
\Statex \textbf{Input:} Dataset D2
\Statex \textbf{Output:} CNN model M2
\For{$epoch = 1$ to $L$}
    \For{$batch = 1$ to $N$}
    \State{Randomly sample a mini-batch training data from D2 and update the weights of input model M2.}       
    \EndFor
\EndFor
\end{algorithmic}
\end{algorithm}

After the model architecture search stage, we use the model with the highest accuracy to perform the final model training in Brevitas. The algorithm used in the model training stage is listed as Algorithm~\ref{a2}. Brevitas is similar to Pytorch, with the added ability to quantize the layers of the model, such that these layers can have quantized input values, weights, and output values. The bit width of the quantized values can be defined by the user. After training, Brevitas outputs the trained network in ONNX format which is used in the next step of hardware implementation in FINN.

\subsection{Hardware implementation}
\label{Hardware implementation}

\begin{figure}
    \centering
    \includegraphics [width=0.6\columnwidth]{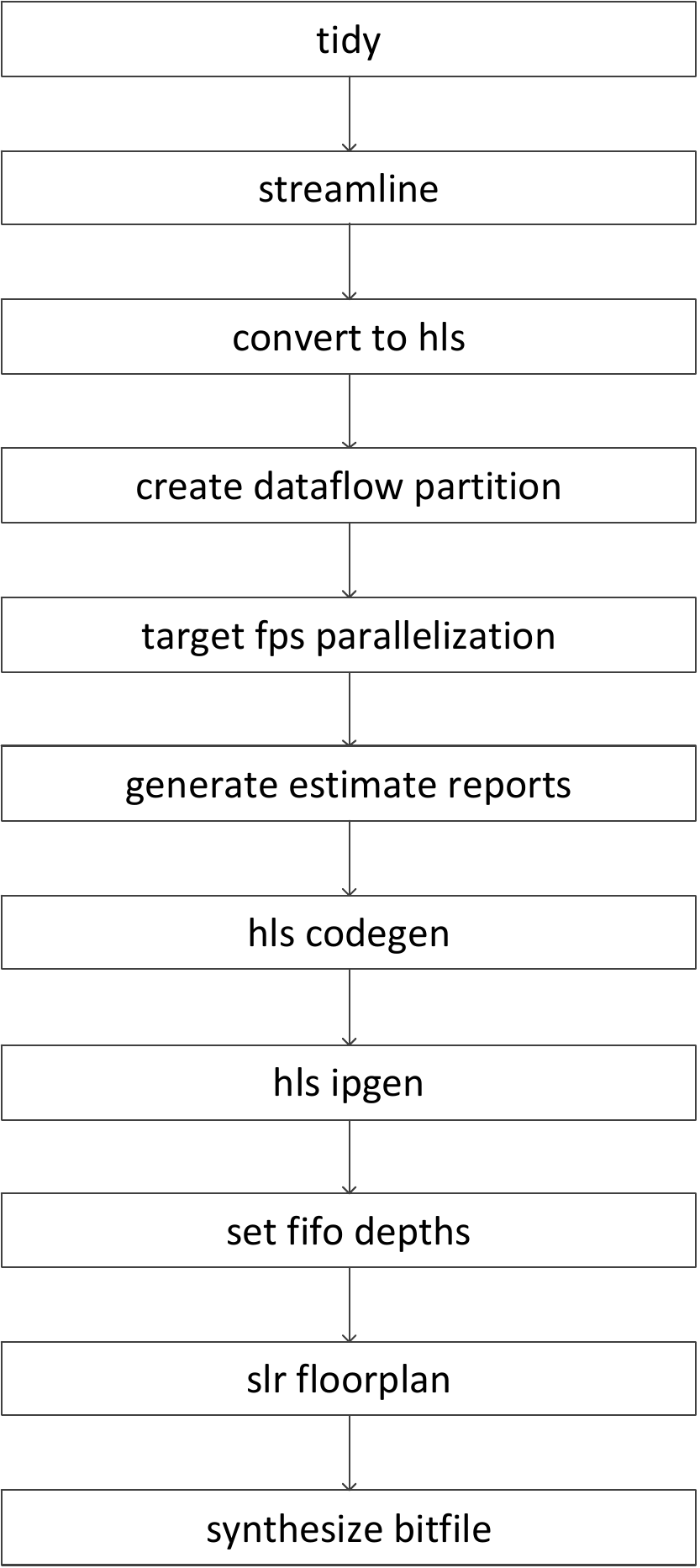}
    \caption{Various steps of the FINN flow}
    \label{fig3.5}
\end{figure}

FINN is the tool we use in order to generate our hardware implementation of our models. The FINN flow we use in this paper is shown in Figure \ref{fig3.5}. FINN uses a number of steps to compile the ONNX model description to hardware, starting with tidy, streamline, convert to hls, etc. Since the NASH model is generated using a special design flow and is different from classic models commonly generated, extra transformation steps are needed to be used. The MoveMulPastMaxPool transform is added in the streamline step since it is needed to process the max pooling layers used in the NAS-convolutional cell (Table~\ref{tb3.1}). We also add the AbsorbSignBiasIntoMultiThreshold transform at the beginning of the streamline step since it is needed by the identity activation function (Figure~\ref{fig3.3}). For other steps, we follow the standard steps in the FINN flow.

\section{Experimental results}
\label{Experimental Results}
This section discusses the experimental results of our NASH models. The results measure various implementation parameters such as accuracy, latency, throughput, and resource utilization. Comparisons with other related networks will also be shown in this section.

\subsection{Accuracy}

We use NASH to optimize the model architecture of ResNet18. We analyze the results of the optimized models with different bit widths of weight and activation. We also implement the four different versions of the NASH as illustrated in Section \ref{Training algorithm}. 
We use the CIFAR-10~\cite{Krizhevsky2009LearningML} data set to search the model architecture, while we use the ImageNet (ILSVRC2012) classification~\cite{WOS:000365089800001} data set to train and evaluate the original and NASH ResNet18 model. The batch size we use for ResNet18 is 256, and we trained 100 epochs for each model. 

\begin{figure}
    \centering
    \includegraphics [width=0.98\columnwidth]{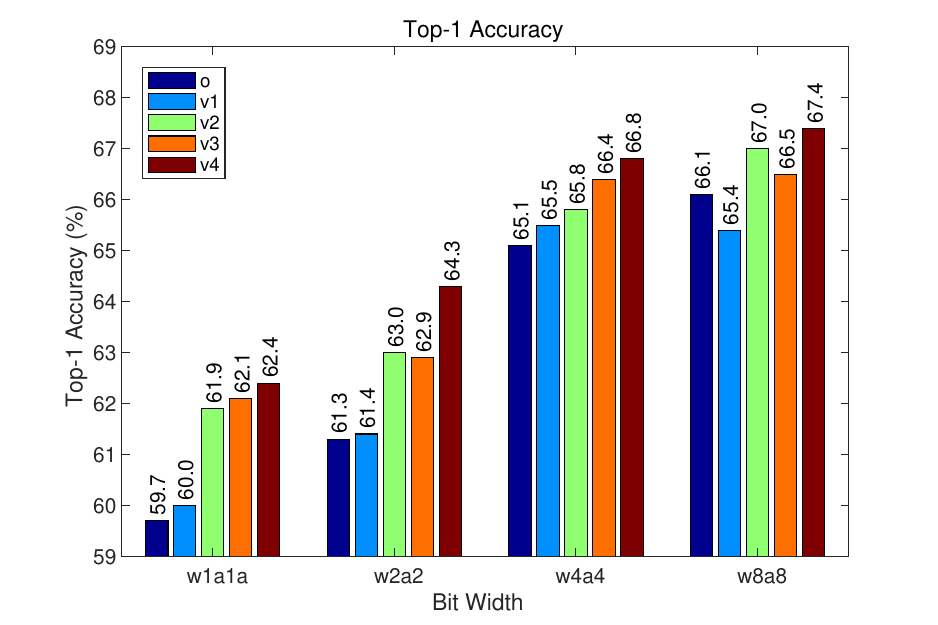}
    \caption{The Top-1 accuracy of ResNet18 and NASH ResNet18 with different bit widths and different NASH versions}
    \label{figtop1}
\end{figure}
~
\begin{figure}
    \centering
    \includegraphics [width=0.98\columnwidth]{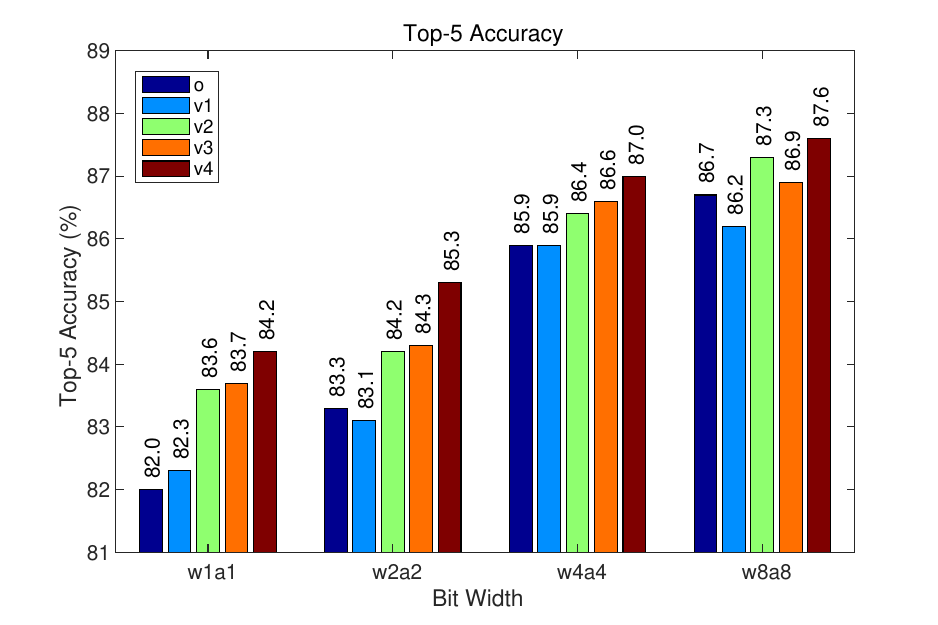}
    \caption{The Top-5 accuracy of ResNet18 and NASH ResNet18 with different bit widths and different NASH versions}
    \label{figtop5}
\end{figure}

Figure \ref{figtop1} and \ref{figtop5} show the results of Top-1 and Top-5 accuracy with different versions of NASH and different bit widths we used for weights and activations.
 
NASH-v1 through NASH-v4 is represented by v1 through v4 in the table, as discussed in Section \ref{Training algorithm}. The "o" in the table refers to the original networks without using the NASH strategy. w1a1 means the weights of the main branch of the ResNet18 model are 1-bit and the activation functions of the main branch are 1-bit (the same interpretation is used for w2a2 through w8a8). For all versions, the bit width of the weights and the activation functions in the residual branches are 8-bit, and the activation function of the NASH branches is 8-bit. For NASH-v1 to v3, the weights of the NASH branches are 1-bit, and for NASH-v4, they are 8-bit. 

Figure \ref{figtop1} and \ref{figtop5} show that, except for some entries of NASH-v1, all the NASH versions have better accuracy than the original version. NASH-v4 has the best accuracy among all NASH versions. The accuracy is 1.3\% to 3.0\% higher in Top-1 accuracy, and 0.9\% to 2.2\% higher in Top-5 accuracy than the original version. The NASH-v1 has the lowest accuracy among all NASH versions, however, it still shows a 0.1\% to 0.4\% increase in Top-1 accuracy in the w1a1 through w4a4 models. The accuracy results of the NASH-v2 and v3 are similar to each other. For the w1a1 and w4a4 models, NASH-v3 has higher accuracy than NASH-v2. In contrast, for the w2a2 and w8a8 models, NASH-v2 has higher accuracy than NASH-v3.

\subsection{Latency, throughput, and resource utilization}

We use FINN 0.9 to generate a hardware implementation of the neural networks. We use the largest parallelism settings in FINN and keep the throughput the same for all versions of the neural networks. Table \ref{tb4.2} lists the results for different versions and different bit widths of the original and NASH ResNet18. These results are calculated by FINN.

Table \ref{tb4.2} shows that the latency and throughput do not depend on the bit width of the networks. The original version of ResNet18 has the lowest latency which is 116.0 ms. The four different NASH versions have almost the same latency, which is about 134.4 ms; this is 15.8\% higher than the original version. This increase can be attributed to the increased number of operations used by NASH to implement the models in HW. The throughput of all the implemented neural networks is 324.54 fps.

\begin{table}[]
\caption{Latency and throughput of ResNet18 and NASH ResNet18 with different bit widths and different NASH version}
\label{tb4.2}
\centering
\begin{tabular}{|c|c|c|c|c|c|}
\hline
bit width & \multicolumn{5}{c|}{w1a1} \\ \hline
version       & o     & v1     & v2    & v3    & v4    \\ \hline
latency (ms)     & 116.0 & 135.0 & 133.7 & 134.6 & 134.1 \\ \hline
throughput (fps) & 324.5 & 324.5 & 324.5 & 324.5 & 324.5 \\ \hline
\multicolumn{6}{c}{ } \\ \hline
bit width & \multicolumn{5}{c|}{w2a2} \\ \hline
version       & o     & v1     & v2    & v3    & v4    \\ \hline
latency (ms)     & 116.0 & 135.0 & 133.7 & 134.4 & 134.1 \\ \hline
throughput (fps) & 324.5 & 324.5 & 324.5 & 324.5 & 324.5 \\ \hline
\multicolumn{6}{c}{ } \\ \hline
bit width & \multicolumn{5}{c|}{w4a4} \\ \hline
version       & o     & v1     & v2    & v3    & v4    \\ \hline
latency (ms)     & 116.0 & 134.0 & 133.7 & 134.4 & 134.1 \\ \hline
throughput (fps) & 324.5 & 324.5 & 324.5 & 324.5 & 324.5 \\ \hline
\multicolumn{6}{c}{ } \\ \hline
bit width & \multicolumn{5}{c|}{w8a8} \\ \hline
version       & o     & v1     & v2    & v3    & v4    \\ \hline
latency (ms)     & 116.0 & 134.0 & 134.0 & 134.4 & 134.1 \\ \hline
throughput (fps) & 324.5 & 324.5 & 324.5 & 324.5 & 324.5 \\ \hline
\end{tabular}
\end{table}

\begin{figure}
    \centering
    \includegraphics [width=0.98\columnwidth]{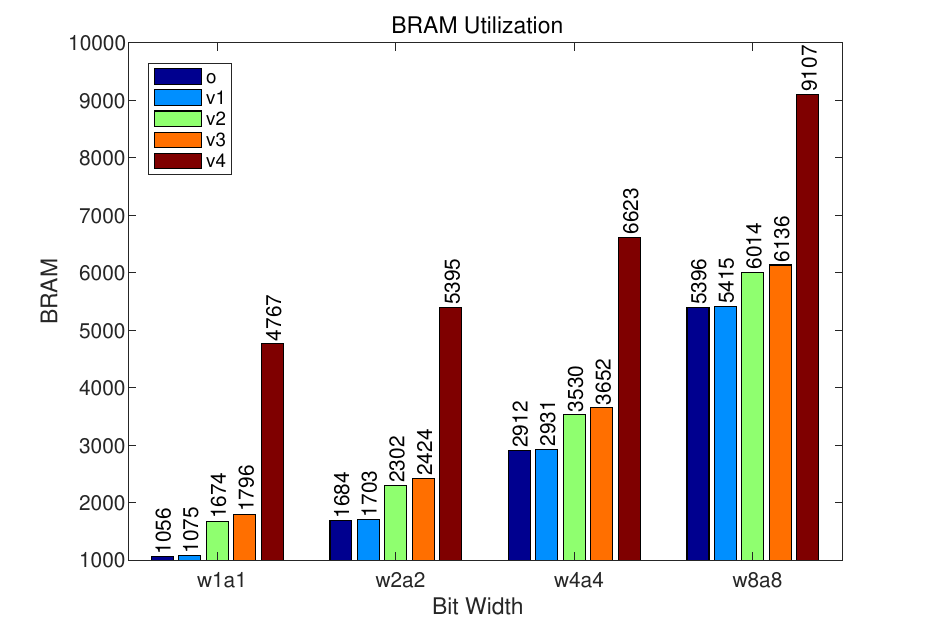}
    \caption{The BRAM utilization of ResNet18 and NASH ResNet18 with different bit widths and different NASH versions}
    \label{figbram}
\end{figure}
~
\begin{figure}
    \centering
    \includegraphics [width=0.98\columnwidth]{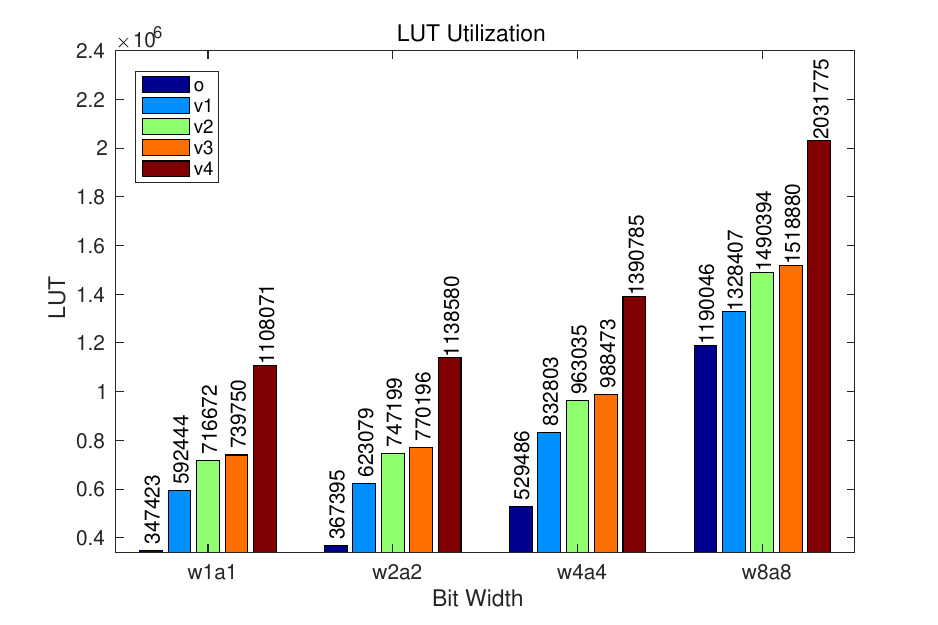}
    \caption{The LUT utilization of ResNet18 and NASH ResNet18 with different bit widths and different NASH versions}
    \label{figlut}
\end{figure}

The resource utilization of different versions and different bit widths of ResNet18 is shown in Figure\ref{figbram} and Figure\ref{figlut}. The resource utilization is based on the latency and throughput results in Table \ref{tb4.2}. In this paper, we use BRAMs and LUTs to compare the results of the original and the NASH versions. The table shows that the original versions have the lowest resource utilization. Among the NASH versions, NASH-v1 has the lowest resource utilization, which is almost the same as the original version, and NASH-v4 has the highest resource utilization. The NASH-v2 and v3 have similar resource utilization, which is about 50\% more than the original version.

\subsection{Trade-off between accuracy and resource utilization}

\begin{figure}
    \centering
    \includegraphics [width=0.98\columnwidth]{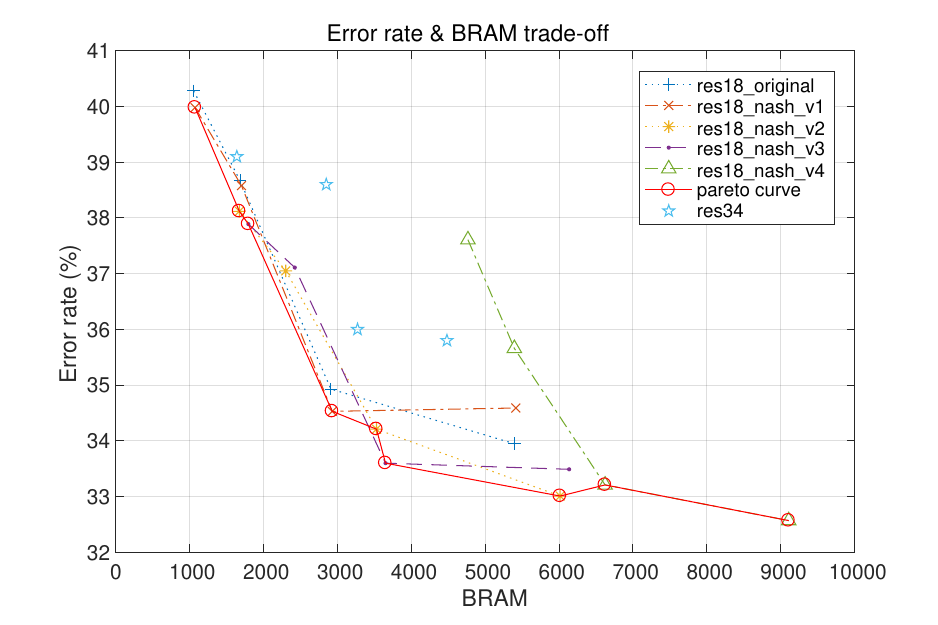}
    \caption{Error rate vs BRAM utilization trade-off}
    \label{fig4.1}
\end{figure}

\begin{figure}
    \centering
    \includegraphics [width=0.98\columnwidth]{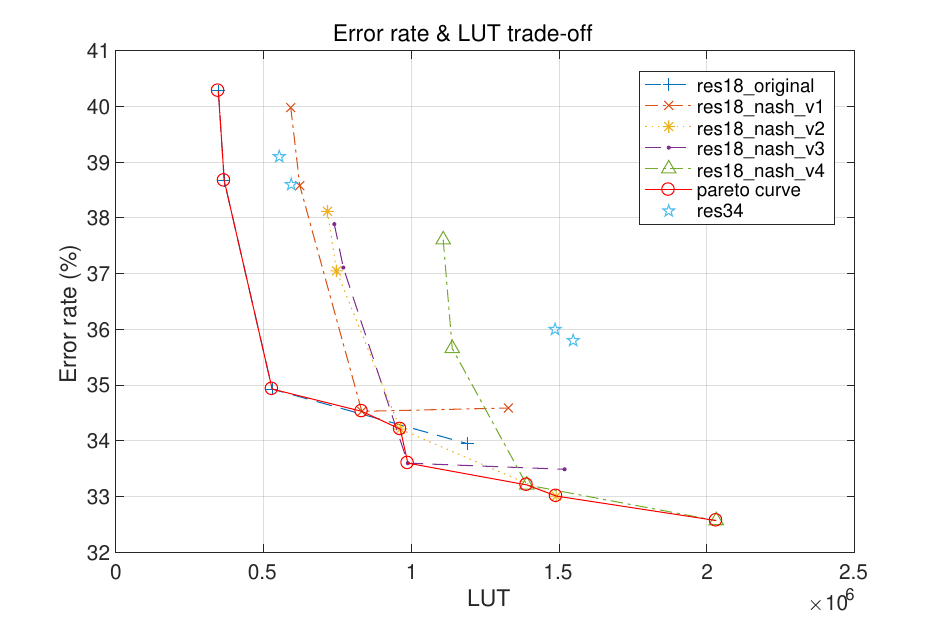}
    \caption{Error rate vs LUT utilization trade-off}
    \label{fig4.2}
\end{figure}

As shown in Figure~\ref{figbram} and Figure~\ref{figlut}, we expect that more HW resources are used as we increase the accuracy of the model, which indicates that there is a trade-off between these two parameters. In order to investigate this trade-off, we plot the relationship between the error rate (calculated as 100-accuracy) and BRAM HW utilization in Figure \ref{fig4.1} as well as the error rate and LUT HW utilization in Figure \ref{fig4.2}. The figures show that we can represent this trade-off as a Pareto curve on the error-HW utilization plane. To find the best balance between accuracy and resource utilization, we need to select the models that have the lowest error for a given HW, or those with the lowest HW for a given error (i.e., models that are close to the bottom-left corner of the plane). 

Figure \ref{fig4.1} shows the trade-off between error rate and BRAM. On each curve, there are four points representing the information of the models with w1a1, w2a2, w4a4 and finally w8a8, respectively. The figure shows that the NASH versions always have a better trade-off than the original version. The models of NASH-v1 with w1a1, w2a2, and w4a4 have a better trade-off than the original models with the same bit width of the weights and activation, where these models of NASH-v1 have a lower error rate and almost the same BRAM utilization. The models of NASH-v2 and NASH-v3 with 1-bit weight and 2-bit activation have a better trade-off compared to the original model with a 2-bit weight and 2-bit activation. The models of NASH-v2 and NASH-v3 with w4a4 and w8a8 have a better trade-off than the original models with the same bit width. NASH-v4 has the best accuracy since it can achieve the highest accuracy among all models, while at the same time demanding the most resource utilization. For applications where the highest accuracy is needed, and there is no limitation on the BRAM utilization, NASH-v4 is the right version to choose.

Figure \ref{fig4.2} shows the trade-off between error rate and LUTs. As shown in the figure, the original models of ResNet18 with w1a1, w2a2, and w4a4 represent the best trade-off for the left side of the figure. The w4a4 models with the NASH-v1, 2, 3, and 4 show a better trade-off between error rate and LUTs than the models with other bit widths. The NASH-v2 model with w8a8 also shows a good trade-off. The NASH-v4 model with w8a8 has the lowest error rate, which is suitable for applications with sufficient resources and high accuracy requirements.

These two figures show that the NASH models have better trade-offs between accuracy and BRAM than the original ResNet18 models with any bit width while having a better trade-off between accuracy and LUTs than the original ResNet18 model with w8a8. In general, FPGAs are usually more limited with BRAM resources than LUTs, making NASH models show good performance in the trade-off of accuracy and resource utilization.

\subsection{Discussion of different NASH versions}

Every NASH version has its benefit in different areas. The models with NASH-v1 have the lowest resource utilization 
which is almost the same as the original models while showing an accuracy increase. NASH-v2 and v3 have a good trade-off between accuracy and hardware resource utilization, they contribute multiple points on the Pareto curves. NASH-v4 is the best accuracy version, it can achieve the lowest error rate with more hardware resource utilization. The models with different NASH versions can be chosen based on different application scenarios.

\subsection{Results of NASH ResNet34}

We also train and implement ResNet34 models with the NASH strategy. We use the CIFAR-10~\cite{Krizhevsky2009LearningML} data set to search the model architecture. In addition, we use the ImageNet (ILSVRC2012) classification~\cite{WOS:000365089800001} data set to train and evaluate the ResNet34 models, which uses a batch size of 128, trained on 100 epochs for each model. The results of accuracy, latency, throughput, and resource consumption are listed in Table~\ref{tbres34}.

\begin{table}[]
\caption{Model accuracy results of ResNet34 for original (o) and NASH-v3 (v3) models}
\label{tbres34}
\centering
\begin{tabular}{|c|c|c|c|c|}
\hline
bit width & \multicolumn{2}{c|}{w1a1} & \multicolumn{2}{c|}{w2a2}\\ \hline
version   & o     & v3     & o    & v3 \\ \hline
Top-1 acc & 60.9\% & 64.0\% & 61.4\% & 64.2\%  \\ \hline
Top-5 acc & 83.0\% & 85.1\% & 83.6\% & 85.4\%  \\ \hline
latency (ms) & 215.74 & 251.67 & 215.74 & 251.67 \\ \hline
throughput (fps) & 324.54 & 324.54 & 324.54 & 324.54 \\ \hline
BRAM & 1640 & 3273 & 2850 &  4483 \\ \hline
LUT  & 553392 & 1486329 & 594004 & 1547781 \\ \hline
\end{tabular}
\end{table}

We implement the ResNet34 with 2 types of bit width: w1a1 and w2a2. Larger bit width assignments will have a high hardware utilization that does not fit on FPGAs. In addition, we only implement NASH-v3 because this version of the NASH models has the best HW vs accuracy trade-off, since it achieves high accuracy without a severe resource utilization increase. NASH-v1 and v2 have a relatively larger HW utilization for the accuracy they achieve, while NASH-v4 has a relatively lower accuracy for the HW utilization needed.

Compared with the original ResNet34 model (indicated as "o" in the table) with w1a1, the NASH ResNet34 model shows a 3.1\% increase in Top-1 accuracy and a 2.1\% increase in Top-5 accuracy. While for the models with w2a2, the NASH ResNet34 model shows a 2.7\% increase in Top-1 accuracy and a 1.8\% increase in Top-5 accuracy, compared to the original ResNet34 model. Compared with the results of the ResNet18 models with the same bit width and strategy, the accuracy of the ResNet34 models has a 0.1\% to 1.9\% increase. Results also show that applying the NASH strategy to the ResNet34 models results in a more obvious accuracy increase than applying the NASH strategy to the ResNet18 models.

The throughput of all original and NASH ResNet34 models is 324.54 fps. The latency of both original ResNet34 models is 215.74 ms, while the latency of both NASH ResNet34 models is 251.67 ms, which is 16.7\% higher than the latency of the original models.

Table~\ref{tbres34} shows that, for w1a1 models, the BRAM utilization of the NASH ResNet34 model is about 2x higher than the BRAM utilization of the original ResNet34 model, while the LUT utilization is about 2.7x higher. For w2a2 NASH ResNet34 models, the BRAM utilization is 1.6x higher while the LUT utilization is 2.6x higher, compared to the original ResNet34 model. Compared with the results of the ResNet18 models with the same bit width and strategy, the resource consumption of the ResNet34 models shows a 53.3\% to 101\% increase.

Results show that applying the NASH strategy to ResNet34 will increase not just the accuracy, but also the latency and resource utilization both compared to the original ResNet34 as well as ResNet18 models. As a result, it is important to check if ResNet34 models achieve a good HW vs accuracy trade-off compared to ResNet18.

To compare the trade-offs, we add the results of the original and NASH ResNet34 models into the trade-off figures, Figure~\ref{fig4.1} and Figure~\ref{fig4.2}. The blue stars with the name "res34" are the results of the original or NASH ResNet34 models. The two stars to the left side of the figure are the results of the original ResNet34 models while the two stars to the right side are the results of the NASH ResNet34 models, in each of the two figures. The two figures show that the result points of the ResNet34 models are located above the Pareto curve we got from the results of the ResNet18 models, which means the ResNet34 models do not represent a better trade-off between resource utilization and error rate than the ResNet18 models. This shows a clear difference between software and HW-based ML models. For software implementations, applying larger and deeper networks will generally result in better accuracy results. However, when it comes to hardware design, considering the trade-off between resource utilization and error rate, larger and deeper networks seem to utilize excessively more HW than smaller networks.

\subsection{Comparison with state-of-the-art networks}

Table \ref{tb4.4} shows the comparison of the accuracy results of ResNet18 of this work and previous works. As shown in the table, our model with the same bit width for the weights (W=2) and activations (A=2) has 1.7\% higher Top-1 accuracy and 0.9\% higher Top-5 accuracy, compared to Dorefa-Net~\cite{zhou2016dorefa}. The NASH-v4 model with the same bit width for the weights (W=1) and activations (A=8) as SYQ~\cite{WOS:000457843604047} has 3.3\% higher Top-1 accuracy and 2.2\% higher Top-5 accuracy than SYQ. Because we do not have models with exactly the same bit width as the NASB model~\cite{WOS:000626021408054}, we use our models with the most similar bit width to compare with the NASB model. We compare our NASH models with 1-bit weights and 1-bit activations with the NASB model~\cite{WOS:000626021408054}. Our model NASH-v2 has 1.4\% higher Top-1 accuracy and 1.4\% higher Top-5 accuracy, NASH-v3 has 1.6\% higher Top-1 accuracy and 1.5\% higher Top-5 accuracy, and NASH-v4 has 1.9\% higher Top-1 accuracy and 2.0\% higher Top-5 accuracy. Compared with the software implementation of the full precision ResNet18 with the highest accuracy, the Top-1 accuracy we can achieve is only 2.2\% lower, and the Top-5 accuracy is only 1.6\% lower.

\begin{table}[]
\caption{Comparisons of NASH with state-of-the-art for ResNet34}
\label{tb4.4}
\centering
\begin{tabular}{|c|c|c|c|c|}
\hline
Model          & W  & A  & Top-1  & Top-5  \\ \hline
Dorefa-Net~\cite{zhou2016dorefa}     & 2  & 2  & 62.6\% & 84.4\% \\ \hline
NASH-v4        & 2  & 2  & 64.3\% & 85.3\% \\ \hline 
\hline
SYQ~\cite{WOS:000457843604047}            & 1  & 8  & 62.9\% & 84.6\% \\ \hline
NASH-v4        & 1  & 8  & 66.2\% & 86.8\% \\ \hline
\hline
NASB~\cite{WOS:000626021408054}           & 1  & 1  & 60.5\% & 82.2\% \\ \hline
NASH-v2        & 1  & 1  & 61.9\% & 83.6\% \\ \hline
NASH-v3        & 1  & 1  & 62.1\% & 83.7\% \\ \hline
NASH-v4        & 1  & 1  & 62.4\% & 84.2\% \\ \hline
\hline
Full precision~\cite{WOS:000638399400004} & 32 & 32 & 69.6\% & 89.2\% \\ \hline
NASH-v4        & 8  & 8  & 67.4\% & 87.6\% \\ \hline
\end{tabular}
\end{table}

\begin{table}[]
\caption{Comparisons of ResNet34}
\label{tb4.5}
\centering
\begin{tabular}{|c|c|c|c|c|}
\hline
Model          & W  & A  & Top-1  & Top-5  \\ \hline
Bi-Real Net~\cite{WOS:000511490100008}    & 1  & 1  & 62.2\% & 83.9\% \\ \hline
NASB~\cite{WOS:000626021408054}           & 1  & 1  & 64.0\% & 84.7\% \\ \hline
NASH-v3        & 1  & 1  & 64.0\% & 85.1\% \\ \hline
\end{tabular}
\end{table}

Table~\ref{tb4.5} compares our NASH ResNet34 model with w1a1 to similar models in literature. Results show that compared with Bi-Real Net, our NASH model shows a 1.8\% increase in Top-1 accuracy and a 1.2\% increase in Top-5 accuracy. Compared with NASB, our NASH model keeps the same Top-1 accuracy and has a 0.4\% increase in Top-5 accuracy.

\section{Conclusion}
\label{conclusion}
In this paper, we present NASH, a neural architecture search strategy to achieve high accuracy for the models implemented on hardware. We integrated the NASH strategy into FINN to automate the process of model generation such that it can be applied to different machine learning models. With the NASH strategy, we show that the ResNet18 models and the ResNet34 models can achieve up to 3.1\% higher accuracy than the models without the NASH strategy. We also construct the Pareto curves of model error vs model HW utilization to investigate the trade-off between accuracy and resource utilization. The results show that in most cases, NASH can achieve the best trade-off on the Pareto curve. Different versions of the NASH strategy show benefits for different applications using various models with different bit-width of weights and/or activation. We also show that smaller models like ResNet18 achieve a better accuracy vs HW trade-off compared to larger models such as ResNet34.

\bibliographystyle{IEEEtran}
\bibliography{main}

\end{document}